\documentclass[10pt,twocolumn,letterpaper]{article}

\usepackage[pagenumbers]{cvpr} % To force page numbers, e.g. for an arXiv version

\newcommand*\diff{\mathop{}\!\mathrm{d}}
\usepackage{algorithm}
\usepackage{algpseudocode}
\algtext*{EndWhile}% Remove "end while" text
\algtext*{EndFor}% Remove "end while" text
\algtext*{EndIf}% Remove "end if" text
\definecolor{cvprblue}{rgb}{0.21,0.49,0.74}
\usepackage[pagebackref,breaklinks,colorlinks,allcolors=cvprblue]{hyperref}

\def\paperID{14} % *** Enter the Paper ID here
\def\confName{CVPR}
\def\confYear{2025}

\title{Fast and Accurate Collision Probability Estimation for Autonomous Vehicles using Adaptive Sigma-Point Sampling}

\author{Charles Champagne Cossette \qquad Taylor Scott Clawson \qquad Andrew Feit\\ \\
Zoox\\
Foster City, CA
}

\begin{document}
\maketitle
\begin{abstract}
A novel algorithm is presented for the estimation of collision probabilities between dynamic objects with uncertain trajectories, where the trajectories are given as a sequence of poses with Gaussian distributions. We propose an adaptive sigma-point sampling scheme, which ultimately produces a fast, simple algorithm capable of estimating the collision probability with a median error of 3.5\%, and a median runtime of 0.21ms, when measured on an Intel Xeon Gold 6226R Processor. Importantly, the algorithm explicitly accounts for the collision probability's temporal dependence, which is often neglected in prior work and otherwise leads to an overestimation of the collision probability. Finally, the method is tested on a diverse set of relevant real-world scenarios, consisting of 400 6-second snippets of autonomous vehicle logs, where the accuracy and latency is rigorously evaluated.

% The ABSTRACT is to be in fully justified italicized text, at the top of the left-hand column, below the author and affiliation information.
% Use the word ``Abstract'' as the title, in 12-point Times, boldface type, centered relative to the column, initially capitalized.
% The abstract is to be in 10-point, single-spaced type.
% Leave two blank lines after the Abstract, then begin the main text.
% Look at previous \confName abstracts to get a feel for style and length.
\end{abstract}

%%%%%%%%% Subfigure example
% \begin{figure*}
%   \centering
%   \begin{subfigure}{0.68\linewidth}
%     \fbox{\rule{0pt}{2in} \rule{.9\linewidth}{0pt}}
%     \caption{An example of a subfigure.}
%     \label{fig:short-a}
%   \end{subfigure}
%   \hfill
%   \begin{subfigure}{0.28\linewidth}
%     \fbox{\rule{0pt}{2in} \rule{.9\linewidth}{0pt}}
%     \caption{Another example of a subfigure.}
%     \label{fig:short-b}
%   \end{subfigure}
%   \caption{Example of a short caption, which should be centered.}
%   \label{fig:short}
% \end{figure*}

\section{Introduction}
\label{sec:intro}
Collision probability estimation between moving agents is a fundamental challenge towards enabling safe motion planning for autonomous vehicles, especially those operating in complex urban environments with real-time constraints. While deterministic collision checking is well understood, incorporating uncertainty in the agent's poses remains a challenging problem. 

The key challenge stems from the fact that the collision probabilities at different times are \emph{dependent} \textemdash the collision probability at a specific time must be conditioned on the previous times being collision-free. The majority of existing methods assume independence \cite{toit2011,tolksdorf2024,park2020,park2017}, which simplifies and accelerates computation, but can lead to overly-conservative estimates. The method in \cite{patil2012estimating} does take this into account, but is intended for a point-mass robot navigating in an obstacle-rich environment. Significant improvements are still required to successfully realize fast, accurate computation of collision probabilities between a self-driving vehicle and another agent within its environment.

\begin{figure}[t]
  \centering
   \includegraphics[width=\linewidth]{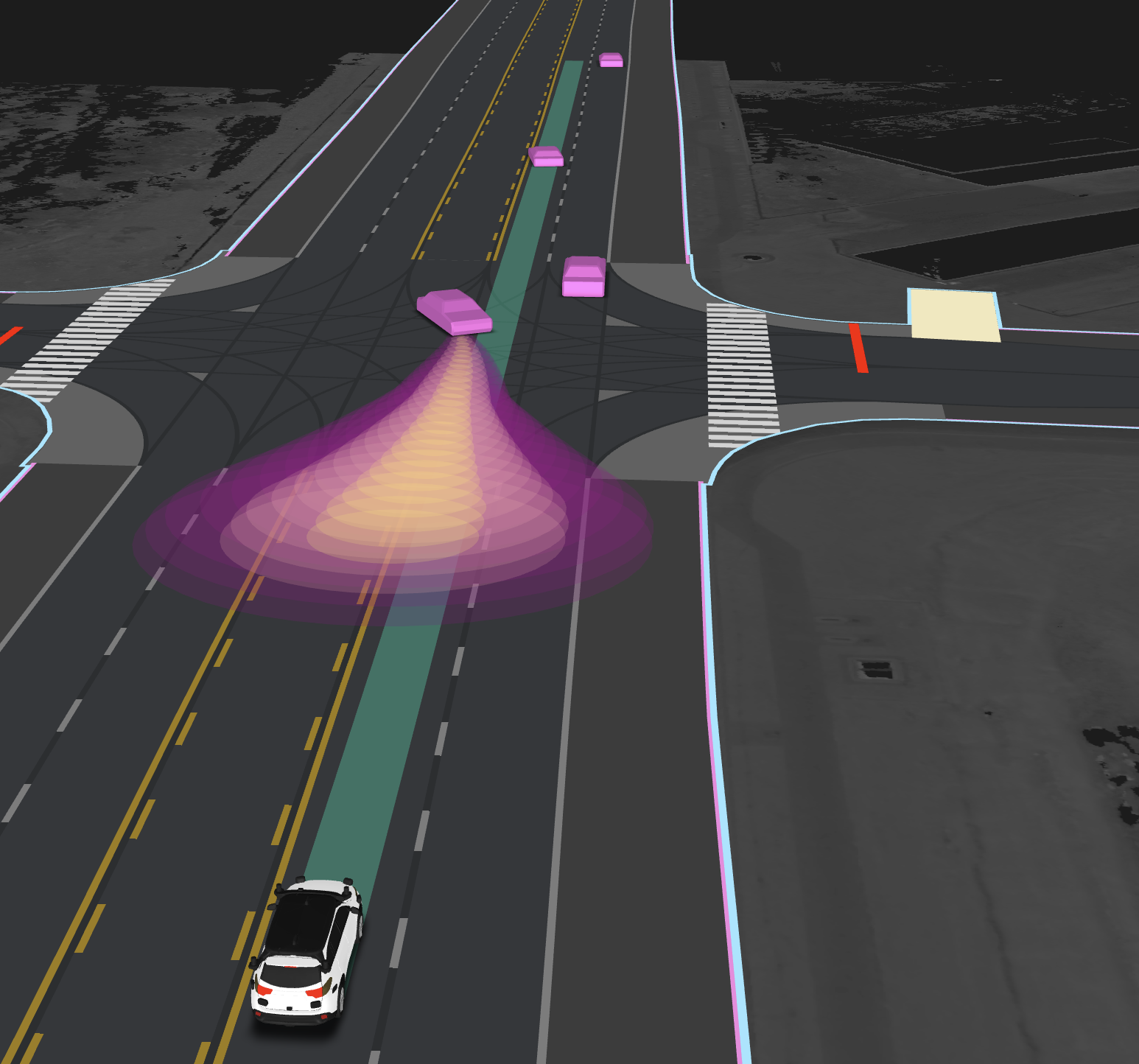}
   \caption{A visualization of a probabilistic trajectory for an agent, consisting of a sequence of Gaussian poses. Computing a collision probability given this prediction, as well as the autonomous vehicle's planned trajectory, is critical for effective collision avoidance.}
   % https://z.zooxlabs.com/argus-staging-20231116T155649-kitt_180-g5oaL
   \label{fig:argus}
\end{figure}

{
\renewcommand\thesubfigure{} % To remove the letter label for this specific figure
\captionsetup[subfigure]{labelformat=empty}
\begin{figure*}
  \centering
  \begin{subfigure}{0.166\linewidth}
    \includegraphics[width=\linewidth]{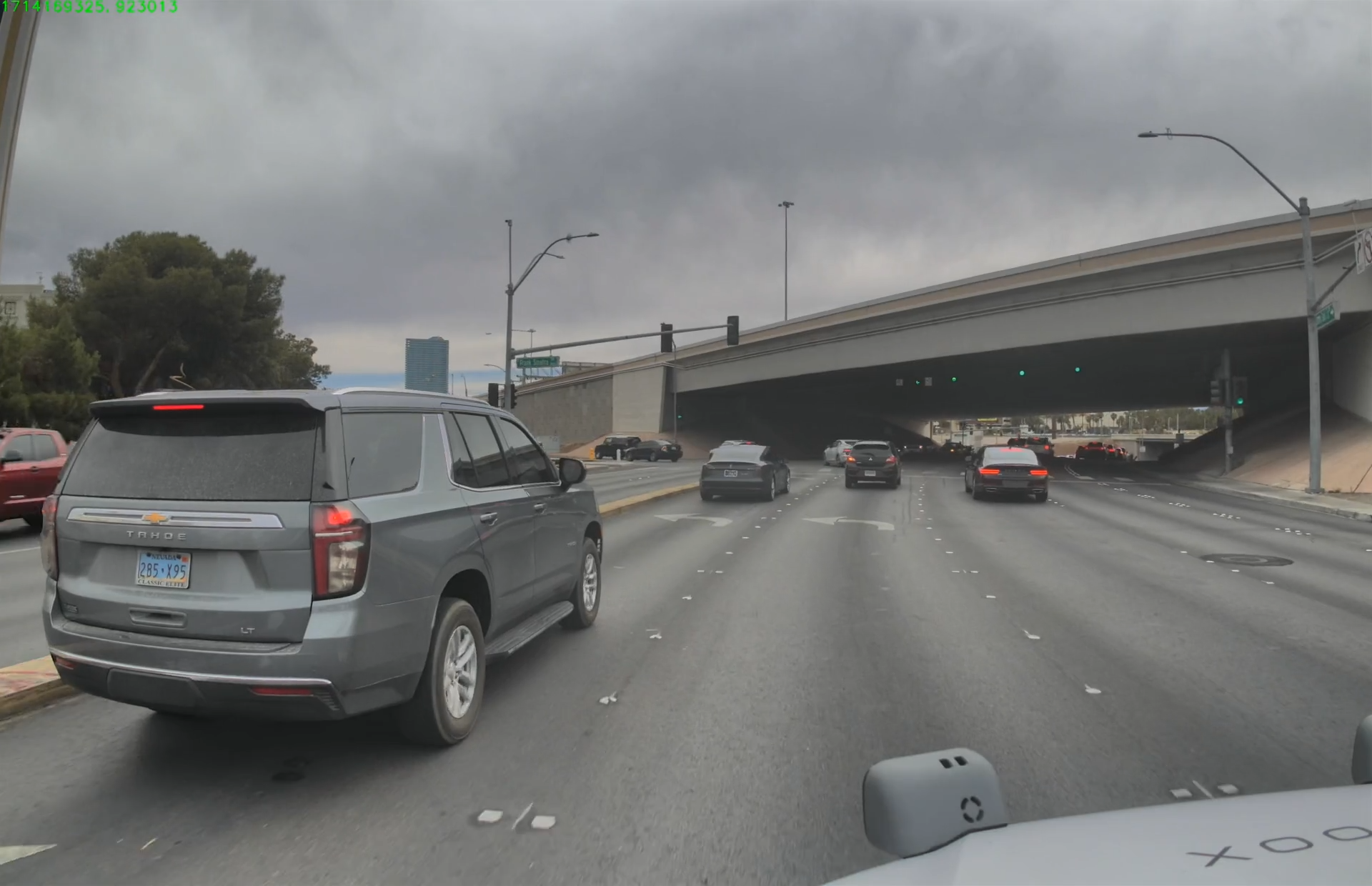}\\
    \includegraphics[width=\linewidth]{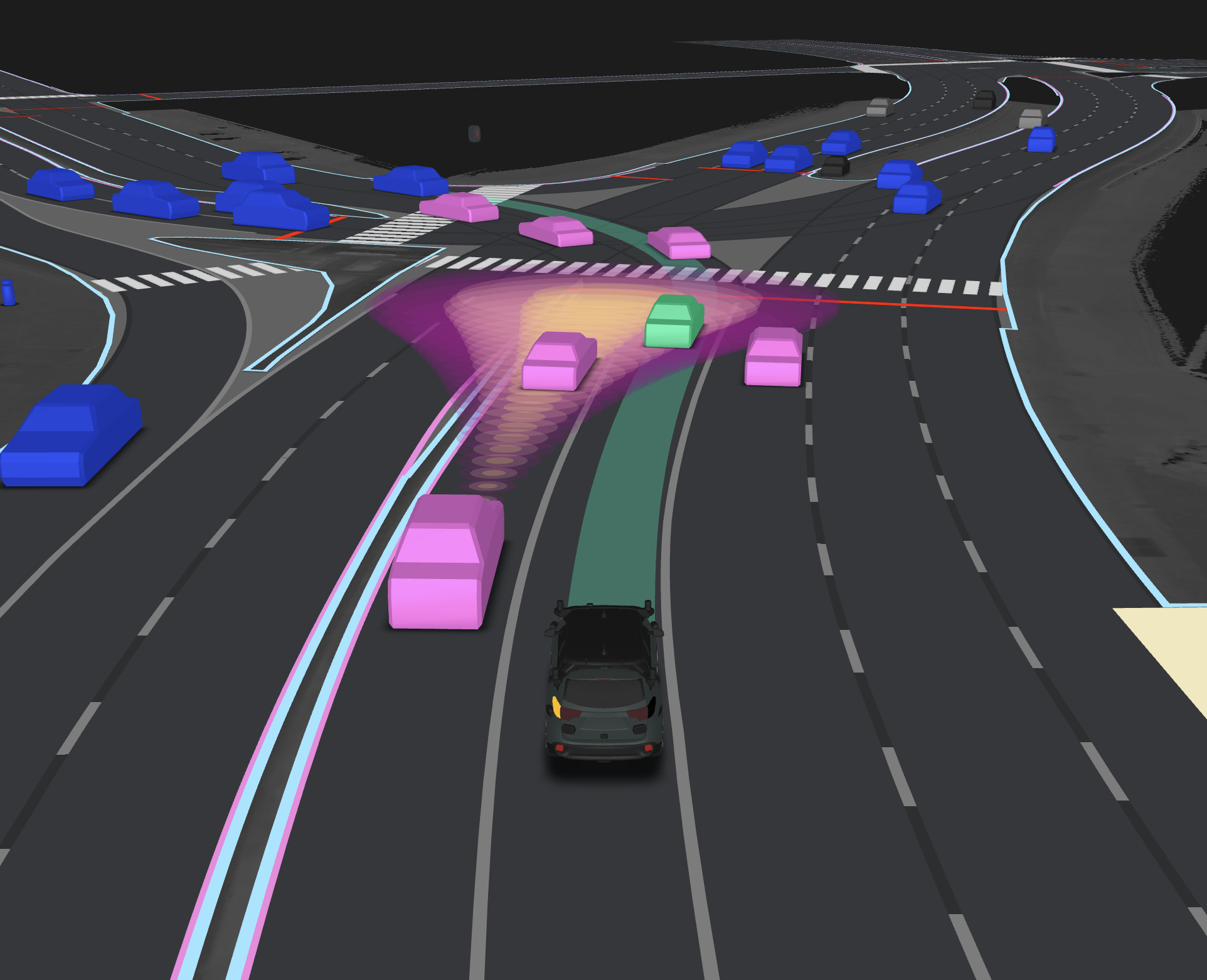}
    \caption{11.56\%}
    \label{fig:short-a}
  \end{subfigure}\hfill
   \begin{subfigure}{0.166\linewidth}
    \includegraphics[width=\linewidth]{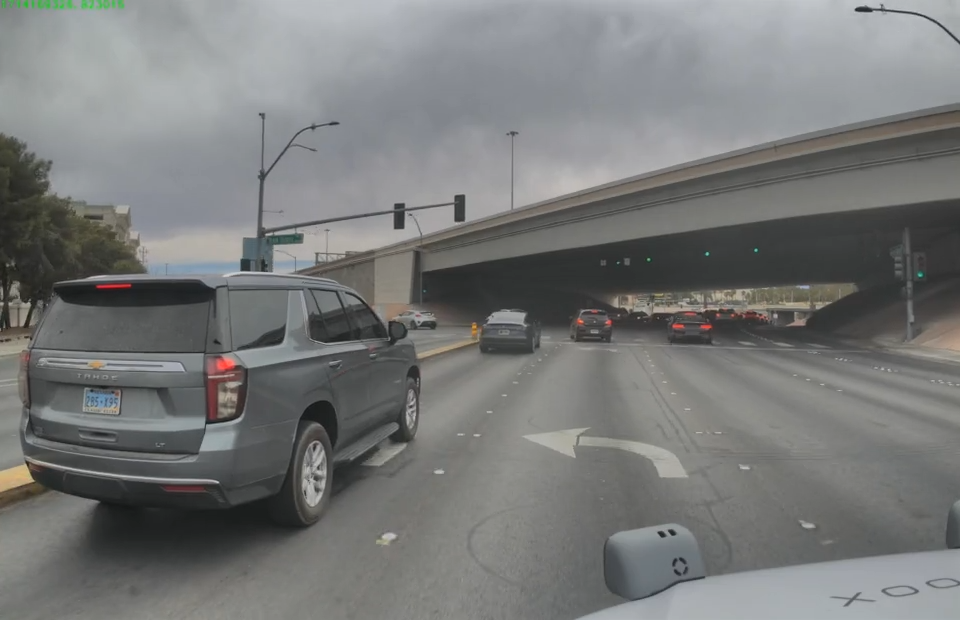}\\
    \includegraphics[width=\linewidth]{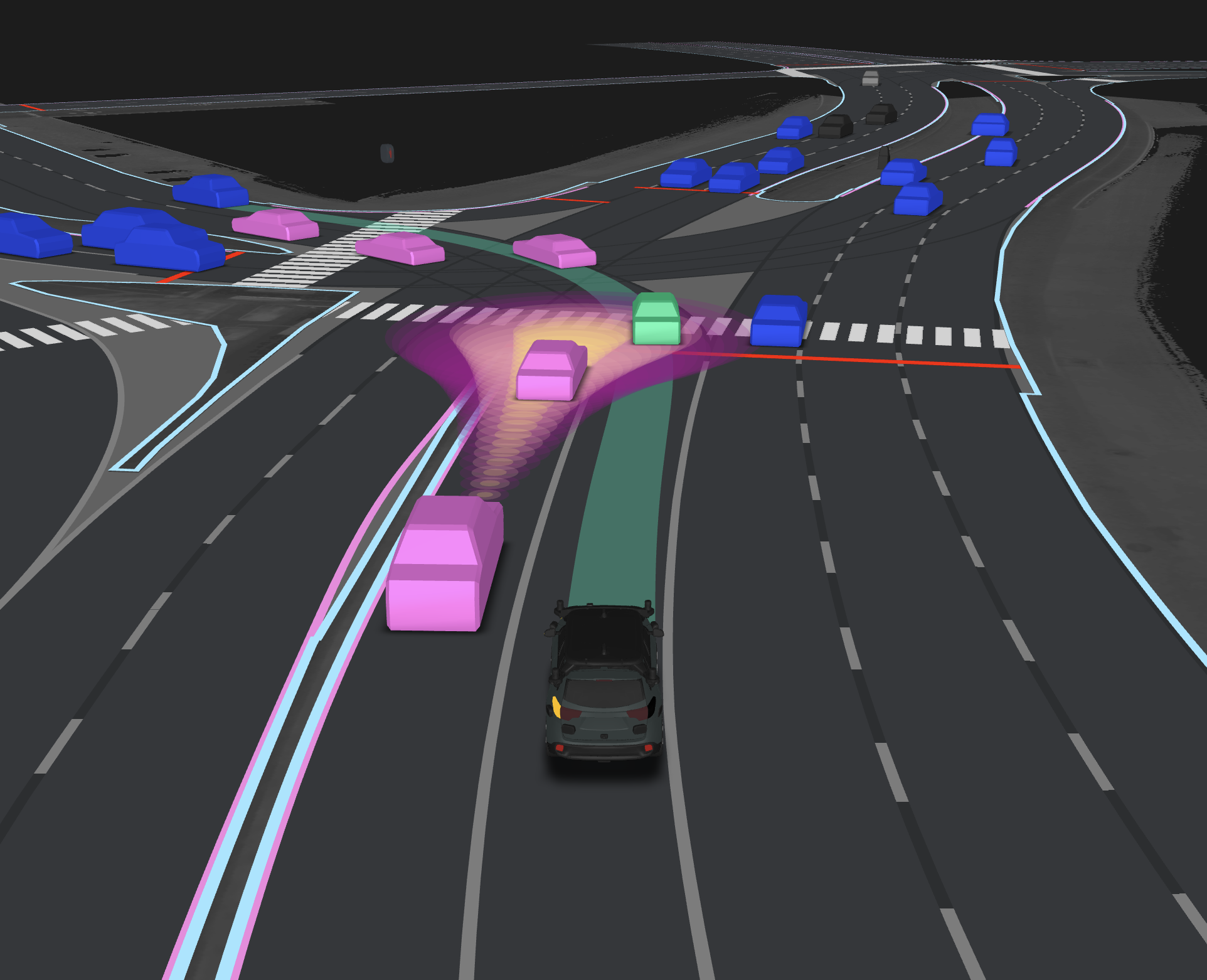}
    \caption{47.00\%}
    \label{fig:short-b}
  \end{subfigure}\hfill
   \begin{subfigure}{0.166\linewidth}
    \includegraphics[width=\linewidth]{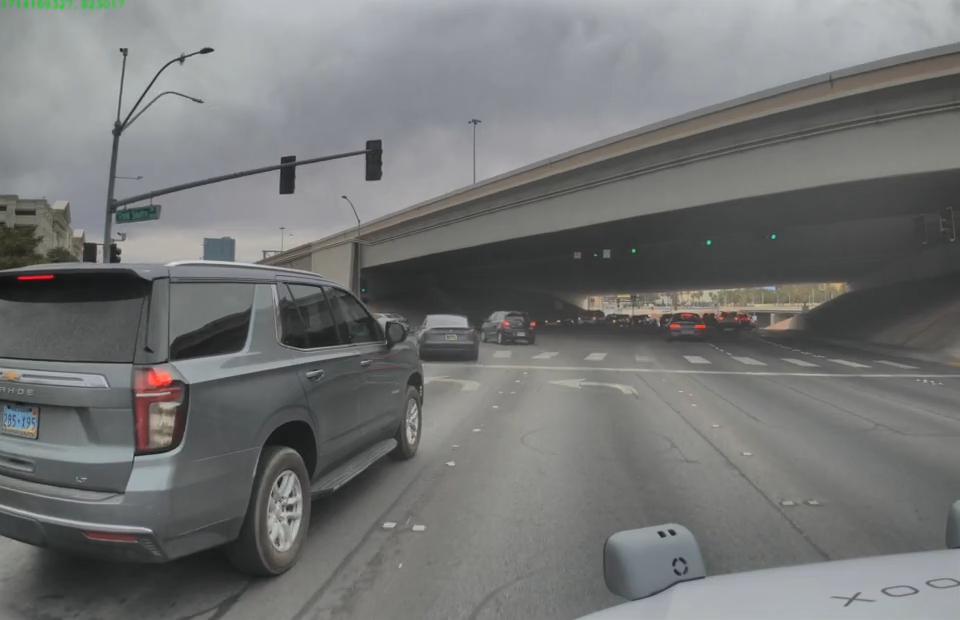}\\
    \includegraphics[width=\linewidth]{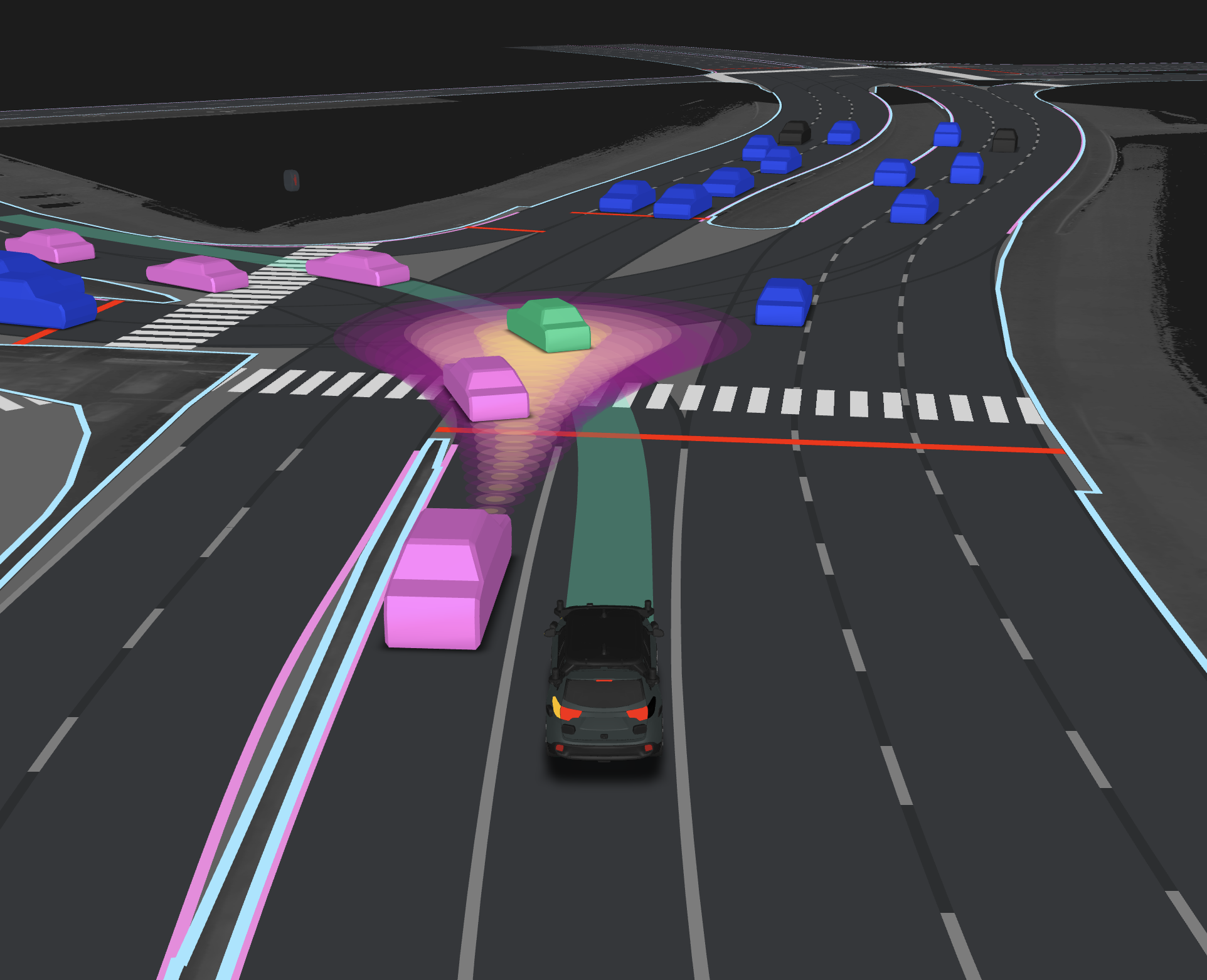}
    \caption{69.94\%}
    \label{fig:short-c}
  \end{subfigure}\hfill
   \begin{subfigure}{0.166\linewidth}
    \includegraphics[width=\linewidth]{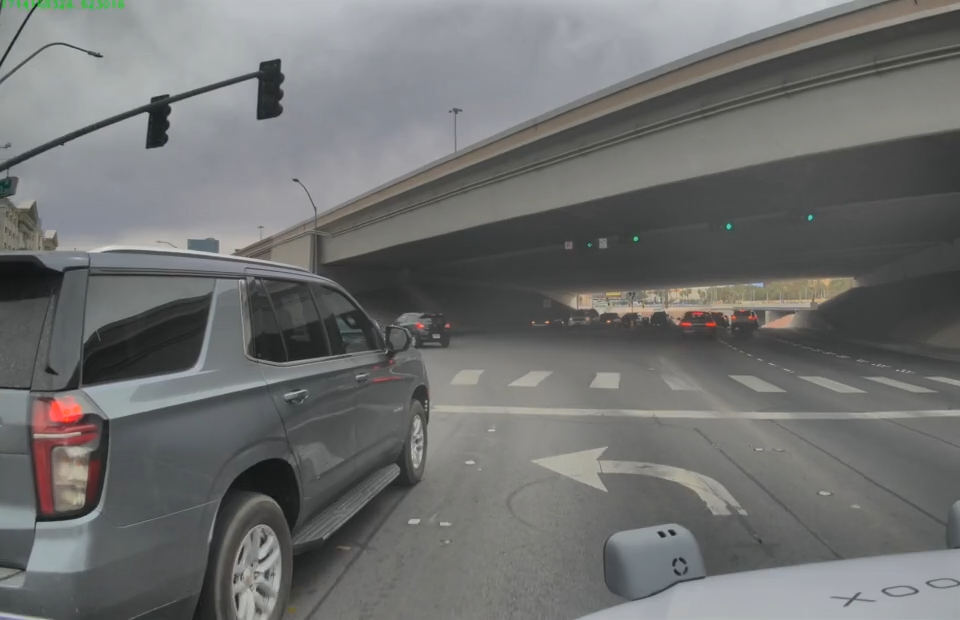}\\
    \includegraphics[width=\linewidth]{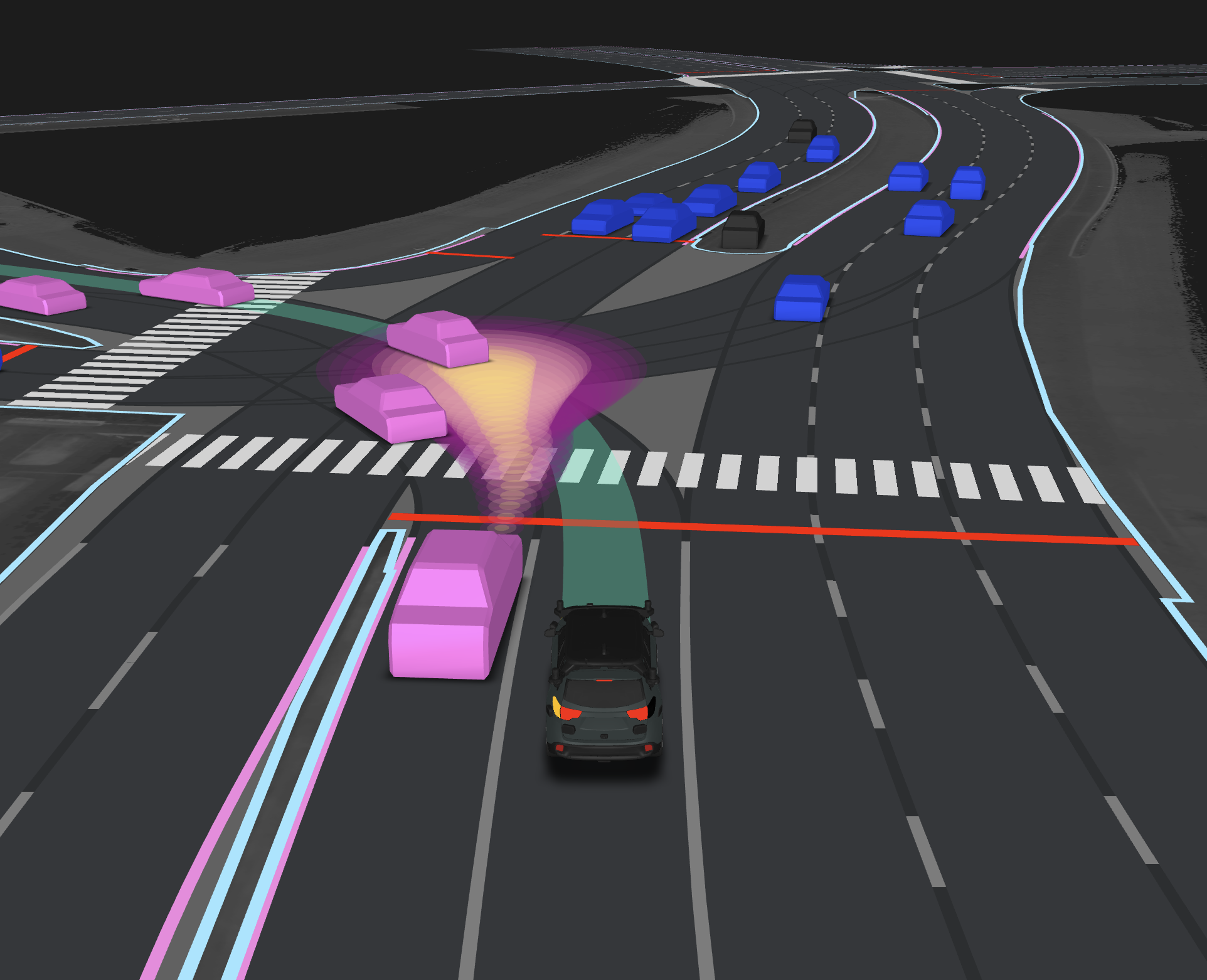}
    \caption{86.29\%}
    \label{fig:short-d}
  \end{subfigure}\hfill
   \begin{subfigure}{0.166\linewidth}
    \includegraphics[width=\linewidth]{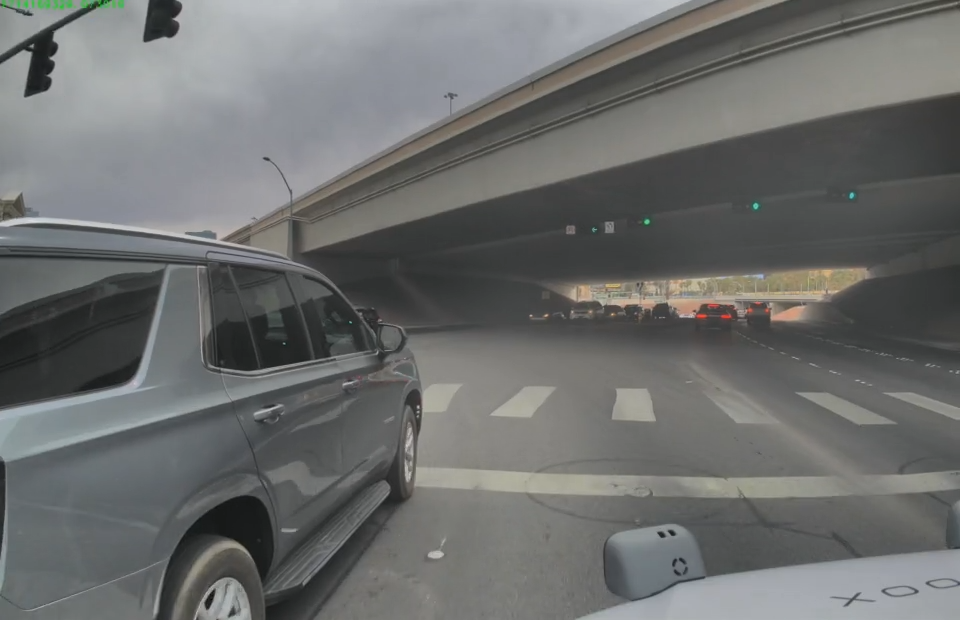}\\
    \includegraphics[width=\linewidth]{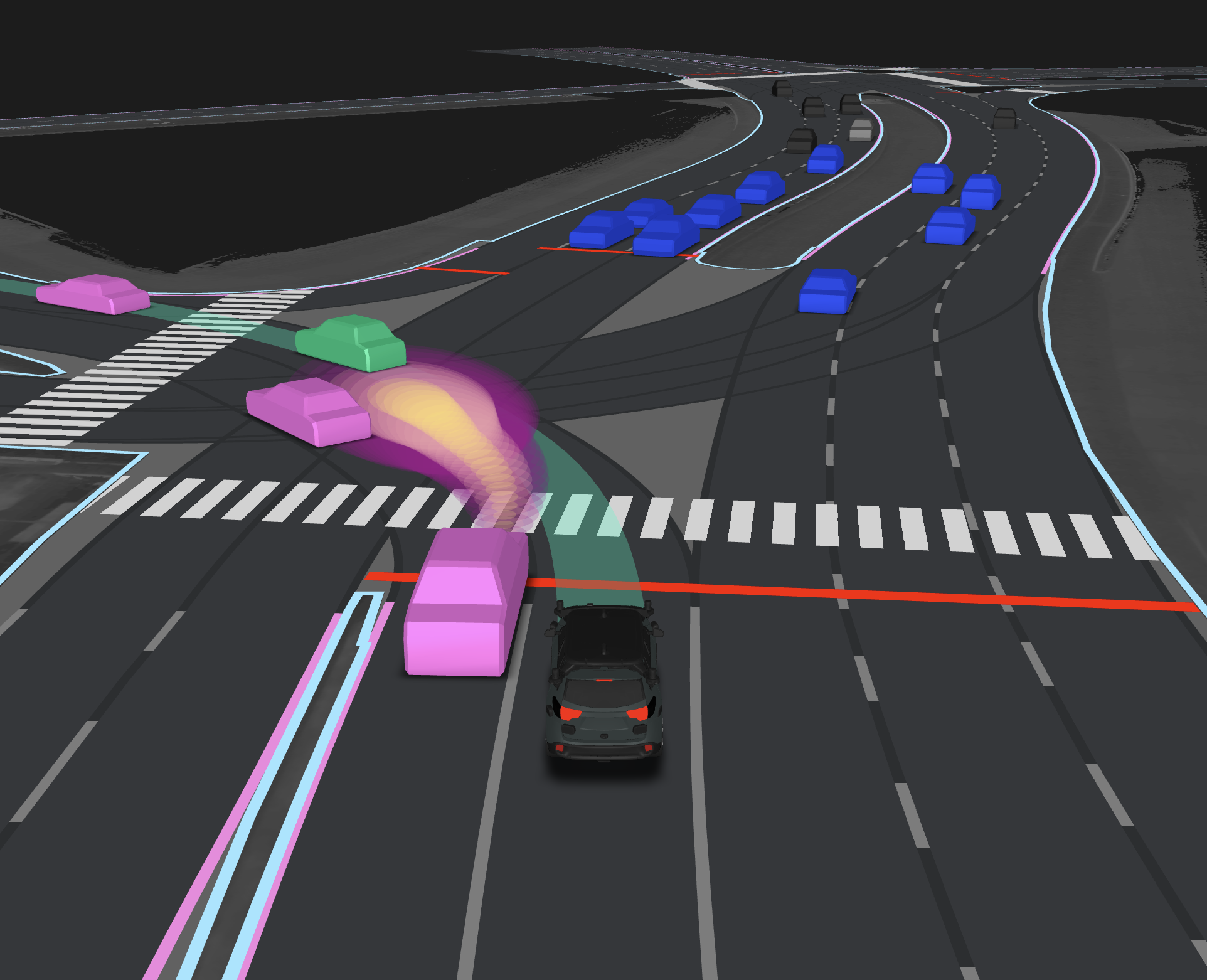}
    \caption{74.04\%}
    \label{fig:short-e}
  \end{subfigure}\hfill
   \begin{subfigure}{0.166\linewidth}
    \includegraphics[width=\linewidth]{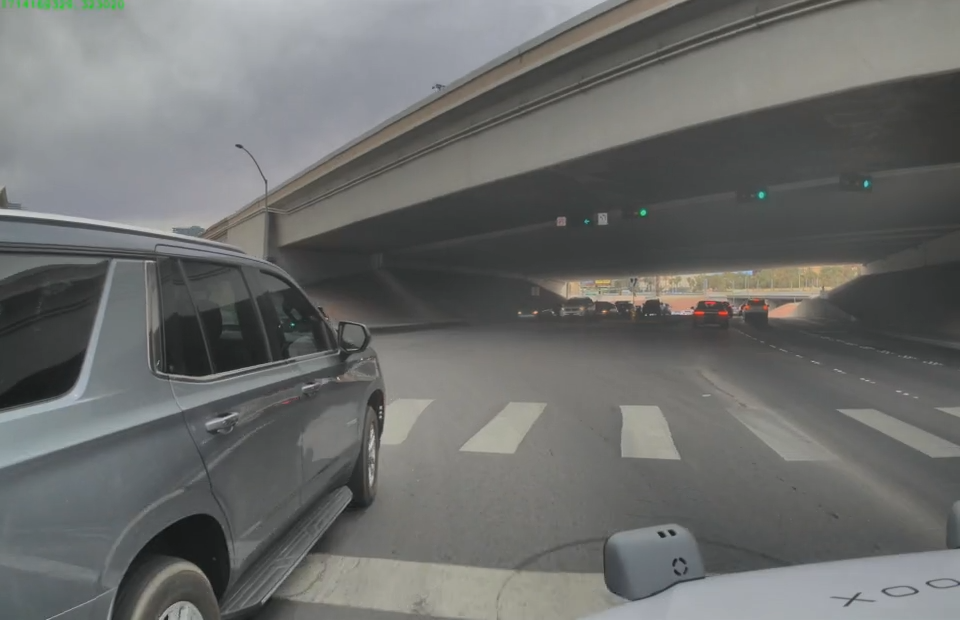}\\
    \includegraphics[width=\linewidth]{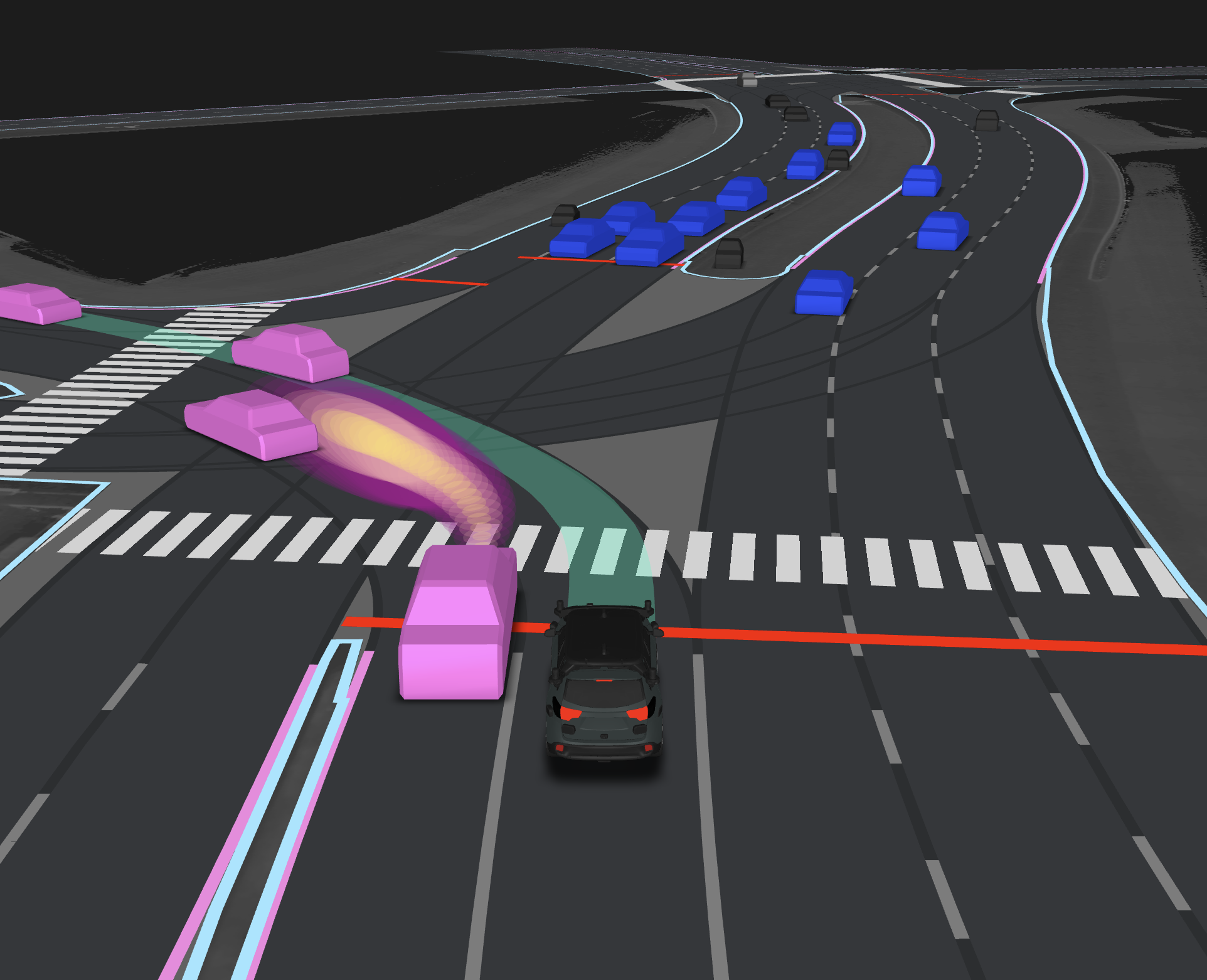}
    \caption{37.88\%}
    \label{fig:short-f}
  \end{subfigure}
  \caption{A real-world, double left-turn scenario with an agent creeping into the autonomous vehicle's lane. The top row consists of frames from a frontal camera at various points in time, whereas the bottom row contains a visualization of that agent's Gaussian prediction, at corresponding points in time. This figure showcases the variability in collision probability over time produced by the proposed algorithm. A reduction in collision probability can be seen in the last two frames, since the agent is eventually predicted to turn left more aggressively.}
  \label{fig:argus_sequence}
\end{figure*}
}

This paper introduces a novel approach to efficiently compute collision probabilities between two agents with non-convex geometries, where their trajectories are given as discrete sequences of poses with Gaussian distributions. The proposed algorithm uses a deterministic, temporally-associated sampling scheme, which allows for incorporating non-convexity in the agent geometries, as well as attitude uncertainty, in a relatively simple way. The computation is accelerated through various optimizations and a novel adaptive sigma-point scheme, which dynamically adjusts sampling density based on the uncertainty magnitude. Our C++ implementation has a median computation time of 0.21ms on a CPU, while consistently presenting less than 10\% error compared to a Monte Carlo-based ground truth.

Concretely, the main contributions include: 
\begin{itemize} 
\item a sampling approach that estimates collision probabilities whilst also incorporating temporal dependence and yaw uncertainty;
\item an adaptive sigma-point scheme that provides accurate probability estimates with minimal samples;
\item efficient geometric optimizations that enable sub-millisecond computation times;
\item empirical validation on both synthetic scenarios and real-world autonomous vehicle data.
\end{itemize}
Figures \ref{fig:argus} and \ref{fig:argus_sequence} show visualizations of probabilistic trajectories associated with various agents in a self-driving context.

\subsection{Literature Review}

Existing collision probability estimation methods can generally be put into one of two categories: closed-form methods and sampling methods.

Closed-form methods often have the advantage of computational efficiency, but unfortunately no known analytical solutions for the collision probability between two arbitrary polygons with Gaussian pose uncertainty. Despite this, closed-form methods can yield reasonable approximations with guaranteed bounds on the over- or under-estimation. 
A method using multi-circular shape approximations is presented in \cite{tolksdorf2024}, providing analytical expressions for collision probabilities with Gaussian uncertainties. While computationally efficient, their method requires approximating vehicle shapes with an array of circles and does not handle temporal dependencies. 
A similar approach is presented in \cite{park2018}, where a high-DOF robot is also represented by an array of spheres, and the authors derive a closed-form approximation that strictly upper-bounds the true collision probability.
Wang et al. \cite{wang2025} propose a collision probability computation technique that involves representing shapes using superquadrics. Their approach leverages a linear chance constraint formulation to provide a closed-form upper bound on collision probability, and furthermore also incorporates both position and attitude uncertainty.
In \cite{park2017}, an efficient algorithm for collision-checking non-convex shapes with Gaussian uncertainty is presented, which uses Gaussian truncation. The authors extend their approach to non-Gaussian error distributions in \cite{park2020} using Gaussian mixture models. However, their approach does not consider temporal correlations between poses in the trajectory. 
The method presented in \cite{patil2012estimating} also uses Gaussian truncation, and successfully incorporates temporal correlations by conditioning the distribution being collision-checked on the previous state being collision-free. However, the method assumes that the collision-free distribution is Gaussian, and hence has deteriorating accuracy when the amount of uncertainty is large. 

Sampling-based methods usually have the advantage of implementation simplicity, as well as flexibility, since any deterministic polygon intersection algorithm can be used for the individual samples. However, sampling-based algorithms face the challenge of computational efficiency, since a naive approach requires a large amount of samples to get reasonable accuracy. A naive Monte Carlo sampling approach is orders of magnitude slower than the state of the art. Nevertheless, improvements have been proposed such as the fast Monte Carlo algorithm in \cite{lambert2008}. Their method achieves millisecond-level computation times by using an optimized sampling approach that reduces complexity from $O(N^2)$ to $O(N)$ while maintaining accuracy within 5\% of ground truth.
Hepeng and Quan \cite{hepeng2020} propose a Finite-Dimensional Monte Carlo (FDMC) method that significantly accelerates collision probability estimation by filtering out low-probability collision points and generating states directly through finite-dimensional distributions rather than propagating them. However, this work also does not address correlations over time.

The proposed method is sampling-based, but contains various novel optimizations that reduce the algorithm runtime to sub-millisecond levels, which is amongst the fastest runtimes we have observed in the literature.
\section{Problem Statement}
\label{sec:problem_statement}
This paper concerns scenarios where there are two agents, with geometries represented by arbitrary, non-convex polygons that have fixed vertices when resolved in their respective body frames. Denote the 2D pose of Agent 1 as $\mathbf{x}^{(1)} = [x^{(1)}\;y^{(1)}\;\theta^{(1)}]^T$, where $x^{(1)},y^{(1)}$ represent the position components and $\theta^{(1)}$ represents the attitude, all relative to some world-fixed frame. A similar definition applies to Agent 2, which has a pose denoted with $\mathbf{x}^{(2)}$.

The agents both have trajectories represented in discrete form as a sequence of $K$ poses, where each pose $\mathbf{x}^{(1)}_{k}$ has an associated Gaussian distribution denoted as
$$ 
\mathbf{x}^{(1)}_{k} \sim \mathcal{N}({\boldsymbol{\mu}}^{(1)}_{k}, \boldsymbol{\Sigma}^{(1)}_{k}), \qquad k=0,\ldots,K-1.
$$
This paper assumes the points on the different agent's trajectories correspond to the same points in time, whether they were initially generated this way, or matched with interpolation.

Define, without agent indices, the relative pose as $\mathbf{x}_k = \mathbf{x}^{(1)}_{k} - \mathbf{x}^{(2)}_{k}$, whose distribution is given by 
$$\mathbf{x}_k \sim \mathcal{N}(\boldsymbol{\mu}_k, \boldsymbol{\Sigma}_k) := \mathcal{N}(\boldsymbol{\mu}^{(1)}_{k} - \boldsymbol{\mu}^{(2)}_{k}, \boldsymbol{\Sigma}^{(1)}_{k}+\boldsymbol{\Sigma}^{(2)}_{k}) .$$
The relative pose is sufficient to determine whether two agents are in collision. For any given time, we define the collision-free set $\mathcal{X}_F$ as the set of all possible relative poses that do not result in a collision. Correspondingly, the collision-free indicator function $I: \mathbb{R}^n \to \{0, 1\}$ is defined as 
$$
I(\mathbf{x}) = \begin{cases} 1 & \text{if} \; \mathbf{x} \in \mathcal{X}_F \\ 0 & \text{otherwise} \end{cases},
$$
meaning it returns $1$ if $\mathbf{x}$ is collision-free, and $0$ otherwise. The indicator function can be implemented using any polygon intersection algorithm. We use the simple ray-casting technique for its ability to handle non-convex geometries. The indicator function can be used to express the probability that $\mathbf{x}$ be collision-free as 
$$ 
P(\mathbf{x} \in \mathcal{X}_F) = \int I(\mathbf{x}) p(\mathbf{x}) \diff \mathbf{x}
$$
where $p(\mathbf{x})$ is the PDF of $\mathbf{x}$.
However, the goal of this paper is to compute the probability that the entire trajectory is collision-free,
\begin{equation}
P\left(\bigwedge_{k=0}^{K-1}\mathbf{x}_k \in \mathcal{X}_F\right) = \prod_{k=0}^{K-1} P\left(\mathbf{x}_k \in \mathcal{X}_F \big| \bigwedge_{i=0}^{k-1}\mathbf{x}_i \in \mathcal{X}_F\right),
\label{eq:full_problem_pdf}
\end{equation}
where a notable challenge arises from the fact that the collision probabilities between timesteps are \emph{dependent}, and hence the collision probability at a particular time must be conditioned on previous times being collision-free.

\section{Solution}
\label{sec:solution}
The key design choice of the proposed solution is to associate relative pose values across time according to their \emph{standardized vectors},
$$ 
\mathbf{z} := \sqrt{\boldsymbol{\Sigma}}^{-1}(\mathbf{x} - \boldsymbol{\mu}),
$$
where $\sqrt{\boldsymbol{\Sigma}}$ refers to the matrix square root of $\boldsymbol{\Sigma}$. Mathematically, arbitrarily using $\mathbf{x}_k$ and $\mathbf{x}_{0}$, this association can be written as
\begin{equation}
    \sqrt{\boldsymbol{\Sigma}_k}^{-1}(\mathbf{x}_k - \boldsymbol{\mu}_k) = 
    \sqrt{\boldsymbol{\Sigma}_{0}}^{-1}(\mathbf{x}_{0} - \boldsymbol{\mu}_{0}), \label{eq:association} 
\end{equation}
which is, again, an equality established \emph{by design}. Rearranging \eqref{eq:association} leads to a deterministic expression for $\mathbf{x}_{k}$ as a function of $\mathbf{x}_{0}$,
\begin{equation}
    \mathbf{x}_k = \sqrt{\boldsymbol{\Sigma}_k}\sqrt{\boldsymbol{\Sigma}_{0}}^{-1}(\mathbf{x}_{0} - \boldsymbol{\mu}_{0}) + \boldsymbol{\mu}_k,
\end{equation}
allowing for all relative poses to be expressed as a function of the first relative pose, $\mathbf{x}_k = f_k(\mathbf{x}_0)$. This dramatically simplifies the expression in \eqref{eq:full_problem_pdf}, which can be shown to reduce to an integral over the single random variable $\mathbf{x}_0$,
\begin{equation}
P\left(\bigwedge_{k=0}^{K-1}\mathbf{x}_k \in \mathcal{X}_F\right) = \int \left(\prod_{k=0}^{K-1} I(\mathbf{x}_k)\right)\mathcal{N}(\boldsymbol{\mu}_0, \boldsymbol{\Sigma}_0) \diff \mathbf{x}_0. \label{eq:full_problem_integral}
\end{equation}
The association by standardized vectors performed in \eqref{eq:association} implicitly assumes that the relative poses between times are fully correlated, and that all randomness stems from the initial distribution of $\mathbf{x}_0$. In general, this is an approximation that may not be realistic depending on how the Gaussian distributions were computed in the first place. Nevertheless, we have observed that the solution performs well in practice, and has the additional advantage of not needing to know the cross-correlations between the poses at all the different times. The proposed algorithm is therefore flexible, and computationally tractable.

\subsection{Sampling-based collision-checking}
Evaluating the integral in \eqref{eq:full_problem_integral} can now be done using a variety of numerical integration techniques. The association by standardized vectors can be exploited by defining a set of $N$ weight-and-sample pairs, 
\begin{equation}
\mathcal{S} = \{(w_0, \mathbf{z}_0)\;,\;\ldots\;,\;(w_{N-1}, \mathbf{z}_{N-1}) \},
\end{equation}
where $\mathbf{z}_n$ represents a standardized-vector sample, and $n$ denotes the sample index. The integral in  \eqref{eq:full_problem_integral} can then be approximated as a weighted sum using 
\begin{multline}
    \int \left(\prod_{k=0}^{K-1} I(\mathbf{x}_k)\right)\mathcal{N}(\boldsymbol{\mu}_0, \boldsymbol{\Sigma}_0) \diff \mathbf{x}_0 \\ \approx \sum_{n=0}^{N-1} w_n \left(\prod_{k=0}^{K-1} I(\sqrt{\boldsymbol{\Sigma}}_k \mathbf{z}_n + \boldsymbol{\mu}_k)\right). \label{eq:weighted_sum_integral}
\end{multline}
By means of the product in \eqref{eq:weighted_sum_integral}, it is sufficient to check, for a particular value of $n$, whether any evaluation of $I(\cdot)$ returns $0$, meaning that that sample is in collision. Therefore, the sampling-based algorithm simply checks whether, for each value of $n$, \emph{all} relative poses across time $\mathbf{x}_k = \sqrt{\boldsymbol{\Sigma}}_k \mathbf{z}_n + \boldsymbol{\mu}_k,\; k=0,\ldots,K-1$ are collision-free, and then sums the weights of all the samples that are collision-free for the entire trajectory. This can be implemented by maintaining a set $\mathcal{S}_F$ representing the collision-free samples, and looping forward through time while removing any samples from that set that result in a collision.
\begin{algorithmic}
\State $\mathcal{S}_F \gets \mathcal{S}$
\For {$k = 0, \ldots, K-1$}
    \ForAll {$(w_n, \mathbf{z}_n) \in \mathcal{S}_F$}
        \If {$I(\sqrt{\boldsymbol{\Sigma}}_k \mathbf{z}_n + \boldsymbol{\mu}_k) = 0$}
            \State $\footnotesize \texttt{// Sample in collision.}$
            \State $\footnotesize \texttt{// Remove from set.}$
            \State $\mathcal{S}_F \gets \mathcal{S}_F \setminus (w_n, \mathbf{z}_n)$
        \EndIf
    \EndFor
\EndFor
\end{algorithmic}

\subsection{Monte Carlo approach}
A simple sampling scheme can be created using a Monte Carlo approach, which consists of defining uniform weights and drawing standardized-vector samples from the standard Gaussian distribution. That is, the elements of $\mathcal{S}$ are given by
\begin{equation}
    w_n = \frac{1}{N}, \quad \mathbf{z}_n \gets \mathcal{N}(\mathbf{0}, \mathbf{1}), \quad n=0,\ldots, N-1, \label{eq:monte_carlo}
\end{equation}
where $\mathbf{1}$ represents the identity matrix. This scheme approaches the true value of the integral in \eqref{eq:weighted_sum_integral} as $N$ becomes large, but can be computationally expensive. We instead propose using an adaptive sigma-point scheme to define the sample set $\mathcal{S}$, which the following section describes.
    
\subsection{Adaptive sigma-points}
We refer to a set of \emph{sigma-points} as a set $\mathcal{S}$ of weighted samples defined through moment-matching, meaning that the weights and standardized-vector samples are chosen to satisfy
\begin{align}
    \sum_{n=0}^{N-1} w_n &= 1, \\
    \sum_{n=0}^{N-1} w_n \mathbf{z}_n &= \mathbf{0}, \\
    \sum_{n=0}^{N-1} w_n\mathbf{z}_n \mathbf{z}_n^T &= \boldsymbol{\Sigma},
\end{align}
and possibly higher-order moments. There exists a variety of sigma-point schemes, such as the Unscented scheme \cite{julier1997unscented} or the Gauss-Hermite scheme \cite{sarkka2023bayesian}. Usage of any existing sigma-point scheme produces a valid collision probability estimator, but with varying accuracy.
\begin{figure}[t]
  \centering
   \includegraphics[width=\linewidth]{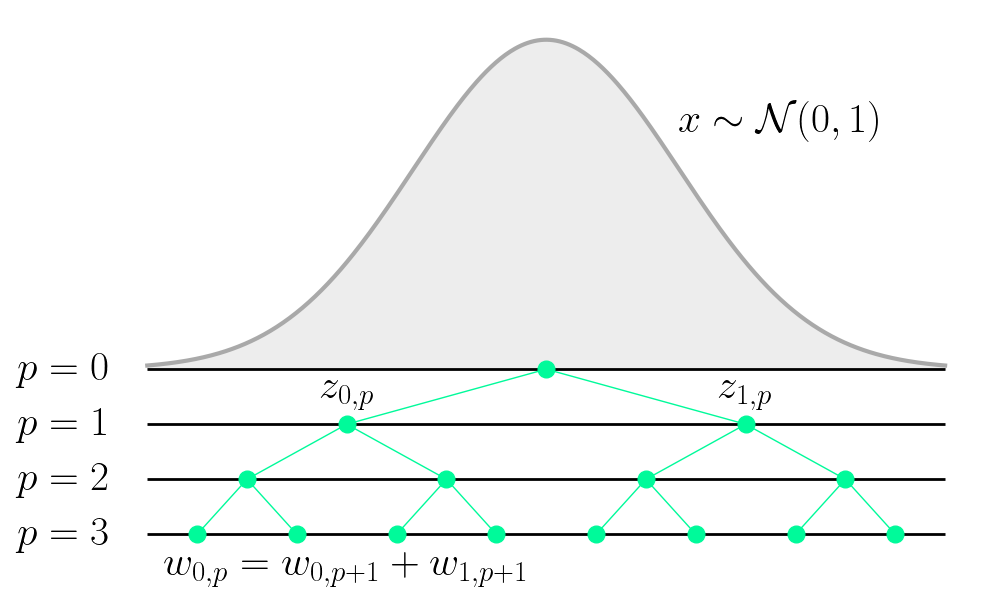}
   \caption{A diagram showing the relationship between sigma-points of varying order, for a 1D distribution.}
   \label{fig:adaptive_sp}
\end{figure}

We have found the popular Unscented scheme to be inaccurate, since for a 2D pose with 3 degrees of freedom, only 7 samples are generated. An 8$^\text{th}$-order Gauss-Hermite scheme with 512 samples performs well, but remains computationally expensive. Hence, we propose an \emph{adaptive} sigma-point scheme, where the quantity of sigma-points changes dynamically with the size of the covariance.

The main insight behind the proposed adaptive scheme is that fewer samples should be used when the norm of the covariance matrix $\boldsymbol{\Sigma}_k$ is small. Considering the temporally-discrete nature of the trajectory, as well as the size of the polygons relative to the covariance norm, generating many samples for a small covariance leads to computing polygon intersection for many nearly-identical polygons. 

The proposed adaptive scheme is illustrated in Figure \ref{fig:adaptive_sp}. Multiple 1D sigma-point sets $\mathcal{S}_p$ are defined for a variety of consecutive orders $p=0,1,\ldots, p_\mathrm{max}$. A coverage interval $\sigma_{max}$, in standard deviations, must also be chosen. The individual 1D sigma-point sets are then obtained by evenly dividing the domain $[-\sigma_\mathrm{max}, \sigma_\mathrm{max}]$ into $2^p$ intervals, placing a sigma-point in each interval center. The $n^\mathrm{th}$ sigma-point value of order $p$ can be shown to be given by 
\begin{equation}
    z_{n,p} = \sigma_{\mathrm{max}} \cdot \left(\frac{2n+1}{2^p} - 1 \right).
\end{equation}
The corresponding weights are chosen to be the probability mass associated with their intervals. That is,
\begin{equation}
    w_{n,p} = \Phi\left(z_{n,p} + \frac{2\sigma_{\mathrm{max}}}{2^p}\right)-\Phi\left(z_{n,p} - \frac{2\sigma_{\mathrm{max}}}{2^p}\right)
\end{equation}
where $\Phi(\cdot)$ corresponds to the cumulative density function associated with the standard Gaussian distribution. 

To form a multi-dimensional sigma-point scheme appropriate for poses, separate 1D sigma-point sets are associated with the $x$- and $y$-components of the relative position, respectively, and with possibly different orders $p_x$, $p_y$. They are combined through the Cartesian product to create a multi-dimensional set corresponding to $\mathbf{x}_k$. That is, the final sigma-point set $\mathcal{S} = S_{p_x} \times S_{p_y}$, corresponding to $x$- and $y$-orders $(p_x, p_y)$, has $2^{p_x} 2^{p_y}$ sigma-points, with weights and sample values given by
\begin{align}
    w_n &= w_{i,p_x} \cdot w_{j, p_y} & i = 0, \ldots, 2^{p_x} -1, \\
    \mathbf{z}_n &= \begin{bmatrix} z_{i, p_x}\; z_{j, p_y} & 0 \end{bmatrix}^T  & j = 0, \ldots, 2^{p_y} -1, \label{eq:pose_sigma_point}
\end{align}
where $n=0,\ldots, 2^{p_x} 2^{p_y}-1$.

While looping through time through the collision checking process, an upsampling step can occur if the covariance size grows sufficiently. When upsampling from order $p$ to order $p+1$, the intervals in order $p$ can be viewed as being split into two sub-intervals of equal width. Ultimately, this results in the fact that
\begin{equation}
    w_{n,p} = w_{2n,p+1} + w_{2n+1, p+1}.
\end{equation}
In other words, for these 1D sets, a particular sigma-point maps to two ``child'' sigma-points, and the child weights of a particular sigma-point sum to the ``parent'' weight. This is the key property that allows upsampling without affecting the collision probability, and upsampling can occur independently along the $x$- or $y$-directions. In our implementation, we choose to upsample along the $x$-direction when 
$$ 
\frac{\Sigma_{xx_k}}{2^{p_x}} > d_{\mathrm{max}},
$$
where $\Sigma_{xx_k}$ is the element in $\boldsymbol{\Sigma}_{k}$ corresponding to the $x$-component of position, and $d_{\mathrm{max}}$ represents the maximum allowable average spacing between sigma-points. An identical condition is used for the $y$-direction. 

The reader should notice that no sigma-points were independently generated for the attitude dimension, and hence the presence of the extra $0$ in the definition of $\mathbf{z}_n$ in \eqref{eq:pose_sigma_point}. This is an intentional design choice, implemented so that the number of sigma-points does not grow too large, for computational reasons. We have observed this to perform well in practice, since the position-attitude correlations present in $\boldsymbol{\Sigma}_k$ are usually large enough to induce varying attitude in the final relative pose samples $\mathbf{x}_k = \sqrt{\boldsymbol{\Sigma}}_k \mathbf{z}_n + \boldsymbol{\mu}_k$. The predictions used in this paper were produced by a Kalman filter operating on kinematic motion models of various vehicles, and hence the constraints present in those kinematics create large position-attitude correlations. Depending on the problem, this assumption may not always hold.

\subsection{Latency optimizations}

\subsubsection{Sigma-point precomputation}
The unit sigma-point values and weights, composing the set $\mathcal{S}$, are fixed and can be precomputed for a variety of $x$- and $y$-orders, and stored in memory. Moreover, we also impose a minimum weight value $w_{\mathrm{min}}$, such that the weight corresponding to any element in $\mathcal{S}_{p}$ is greater than $w_{\mathrm{min}}$. We enforce this in the precomputation step by halting the upsampling for specific sigma-points, if the act of upsampling would produce child sigma-points with weight below $w_{\mathrm{min}}$. In this paper, a maximum order of $p=4$ is used.
 
\subsubsection{Distance checks}
To further accelerate the algorithm, we perform simple radius checks before checking the intersection between the polygons of each sigma-point. Let $r_1$ and $r_2$ be radii that fully contain the polygons of Agent $1$ and Agent $2$, respectively. 

\begin{figure}[t]
  \centering
   \includegraphics[width=\linewidth]{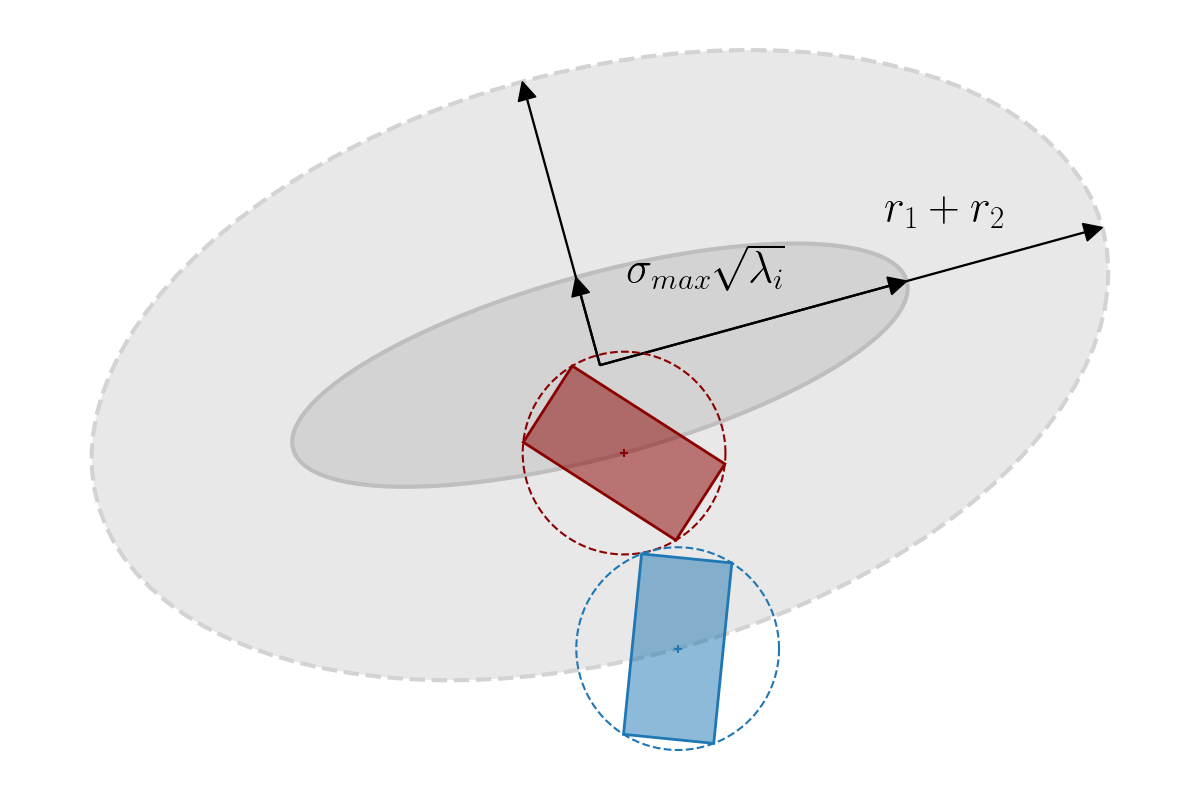}
   \caption{A diagram showing how the covariance ellipse can be inflated to account for the radii of both agents.}
   \label{fig:ellipse_check}
\end{figure}

The first check is to verify whether the polygons are even within the level-set ellipse of $\mathcal{N}(\boldsymbol{\mu}_k, \boldsymbol{\Sigma}_k)$ corresponding to $\sigma_\mathrm{max}$ standard deviations. If not, the entire sampling procedure can be skipped since all sigma-points are guaranteed to be collision-free. This check is visualized in Figure \ref{fig:ellipse_check}.

Let $\mathbf{p}_k \in \mathbb{R}^2$ be the position components of $\boldsymbol{\mu}_k$, and $\boldsymbol{\Sigma}_{p,k}$ be the $2\times2$ sub-block of $\boldsymbol{\Sigma}_k$ corresponding to those position components. This ellipse check is performed by inflating the ellipse corresponding to $\sigma_{max}$ to also account for the agent radii $r_1, r_2$. Denote the eigendecomposition of $\boldsymbol{\Sigma}_{p,k}$ as
\begin{equation}
    \boldsymbol{\Sigma}_{p,k} = \mathbf{V} \; \boldsymbol{\Lambda} \; \mathbf{V}^T. \\ 
\end{equation}
where $\boldsymbol{\Lambda}= \mathrm{diag}(\lambda_1, \lambda_2)$ is a diagonal matrix of eigenvalues. The square-root of the inflated covariance $\boldsymbol{\Sigma}_{\mathrm{max}}$ is given by 
\begin{equation}
    \sqrt{\boldsymbol{\Sigma}_{\mathrm{max}}} = \mathbf{V} \; \left(\sigma_\mathrm{max}\sqrt{\boldsymbol{\Lambda}} + (r_1 + r_2) \mathbf{1}\right) \mathbf{V}^T,
\end{equation}
and all sigma-points are guaranteed to be collision-free if 
\begin{equation}
    || \sqrt{\boldsymbol{\Sigma}_{\mathrm{max}}}^{-1} \mathbf{p}_k || > 1. \label{eq:ellipse_check}
\end{equation}

If the polygons do intersect the ellipse by failing to satisfy \eqref{eq:ellipse_check}, then a simple distance check can be performed for each sigma-point. The $n^\mathrm{th}$ sigma-point at timestep $k$ is guaranteed to be collision-free if
\begin{equation}
    ||\begin{bmatrix}\mathbf{1} & \mathbf{0} \end{bmatrix}(\sqrt{\boldsymbol{\Sigma}}_k \mathbf{z}_n + \boldsymbol{\mu}_k)|| > r_1 + r_2,
\end{equation}
where the matrix $\begin{bmatrix}\mathbf{1} & \mathbf{0} \end{bmatrix}$ simply extracts the position component from the sigma-point.

\begin{figure*}[t]
  \centering
   \includegraphics[width=\linewidth]{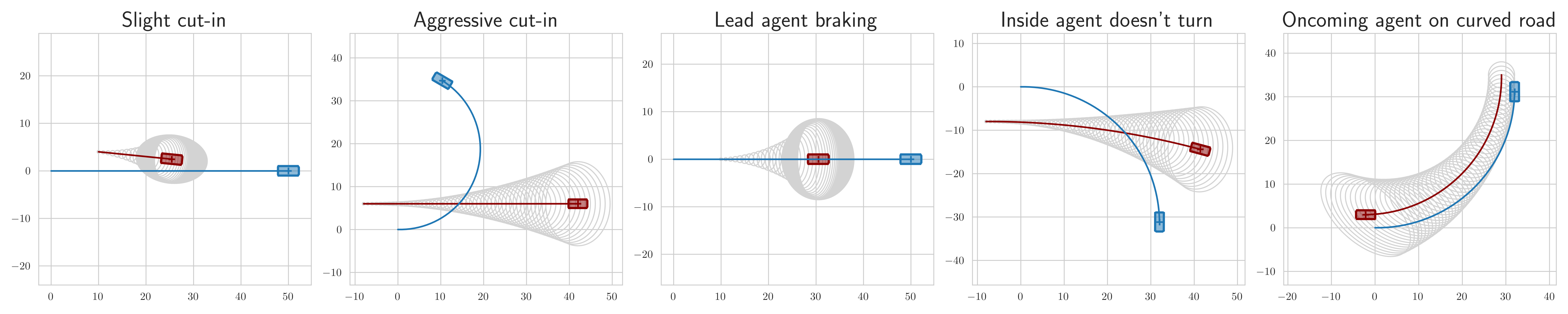}
   \includegraphics[width=\linewidth]{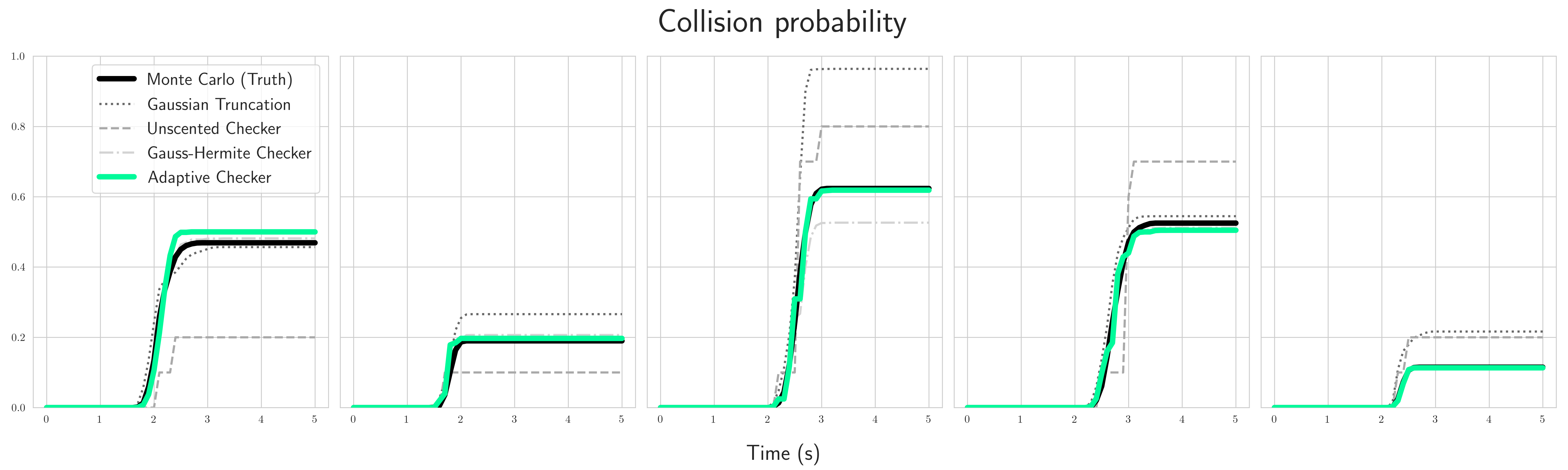}
   \includegraphics[width=\linewidth]{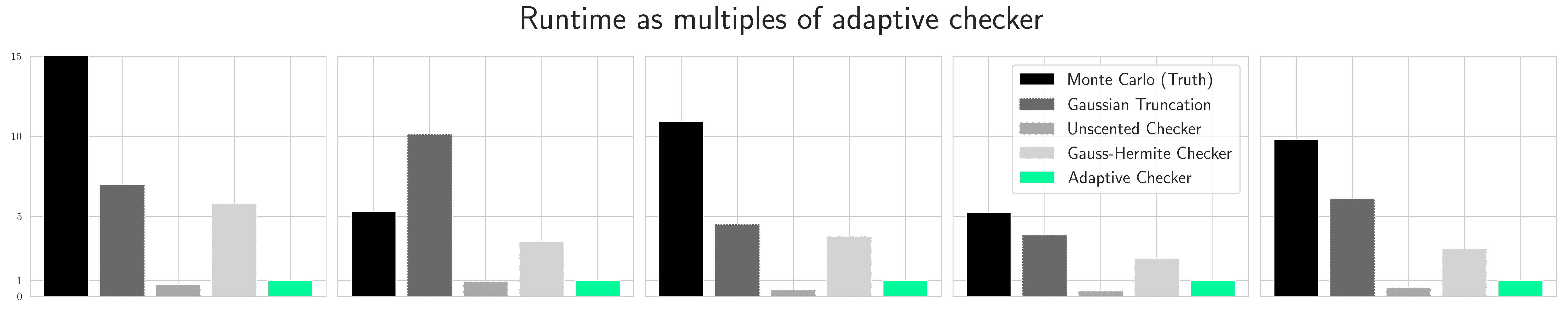}
   \caption{Collision probability and runtime, for a variety of algorithms, for a variety of scenarios. Each column corresponds to the scenario visualized in the top row. The proposed adaptive checker can be seen to closely track the Monte Carlo collision probability throughout the trajectory, yet have a latency comparable to the Unscented checker, which is a fast but inaccurate algorithm using only 7 samples.}
   \label{fig:sim_prob_curve}
\end{figure*}

\subsection{Multiple agents}
The proposed algorithm assumes independence of the pose distributions between agents. Hence, computing collision probabilities for multiple agents is done by evaluating the same algorithm on each agent.

\section{Results}
\label{sec:results}
\subsection{Simulation}
The proposed algorithm was first prototyped in Python, and benchmarked against several alternatives on basic synthetic scenarios. The Monte Carlo-based algorithm described in \eqref{eq:monte_carlo}, with $N=2000$ samples, is used as the source of ground-truth probability values. Other comparison algorithms are formed by using the Unscented and Gauss-Hermite sigma-point schemes \cite{julier1997unscented, sarkka2023bayesian}, which have 7 and 512 samples, respectively. Finally, the Gaussian truncation-based algorithm from \cite{patil2012estimating} is also used as a comparison. This approach works by continuously tracking the distribution of collision-free states throughout the trajectory, and assuming it to be Gaussian.

Figure \ref{fig:sim_prob_curve} shows the collision probability throughout the trajectory, as determined by the various comparisons, and our proposed algorithm labelled as the ``Adaptive Checker''. Figure \ref{fig:sim_prob_curve} also compares the runtime of the same algorithms when executed offline on a 2023 Macbook Pro with the M2 Pro chip. The proposed adaptive algorithm is significantly more accurate than the Gaussian truncation and Unscented algorithms, yet demonstrates a runtime comparable to the fastest algorithm amongst the comparisons, being the Unscented checker. The Unscented checker has a low number of samples, and hence does not have the resolution required to estimate the collision probability accurately. The Gaussian truncation algorithm \cite{patil2012estimating} assumes that the distribution of collision-free states at any point in time is Gaussian, which is an assumption that does not hold well for dynamic agents in self-driving scenarios.

\subsection{Real-world self-driving data}
The adaptive sigma-point collision checker is then implemented in C++ for on-vehicle software, and re-simulated on 400 relevant 6-second snippets of real-world vehicle logs from Zoox's self-driving fleet. The logs contain lists of detected objects in the environment, from which probabilistic predictions are generated, providing estimates for future poses of each agent with corresponding Gaussian uncertainties. These predictions where then used to compute collision probabilities with the proposed algorithm. 

Trajectories with zero collision probability are omitted from the results shown in this section. As an example, Figure \ref{fig:argus_sequence} shows a sequence of frontal camera frames, as well as corresponding visualizations and collision probabilities, for an agent creeping into the autonomous vehicle's lane.

\begin{figure}[t]
  \centering
   \includegraphics[width=\linewidth]{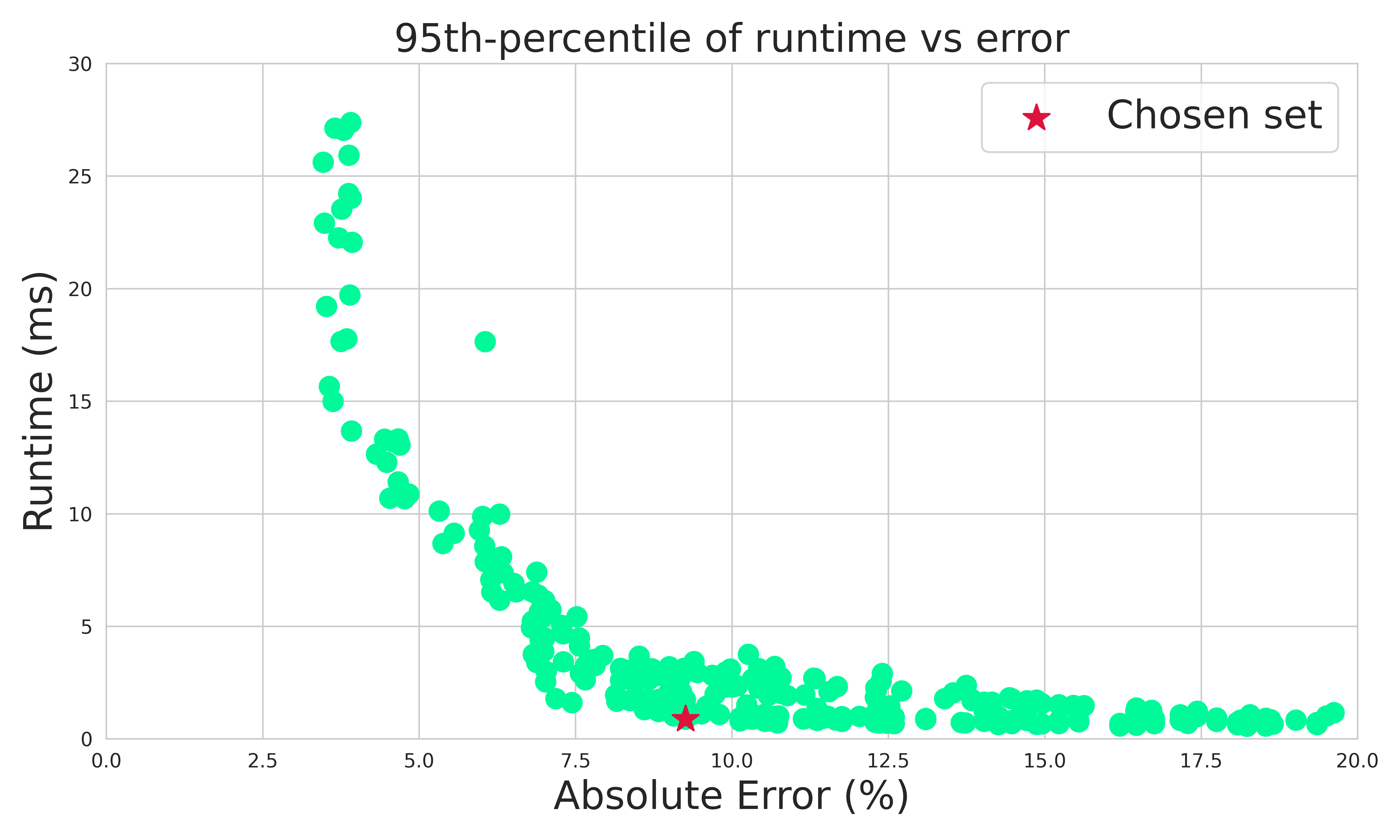}
   \caption{95th-percentile of the algorithm runtime vs error relative to the Monte Carlo collision checker, for a variety of unique parameter choices. The overall curve shows a clear tradeoff between runtime and error.}
   \label{fig:runtime_vs_error}
\end{figure}

This real-world data is used to tune the algorithm's three main tuning parameters:
\begin{itemize}
    \item $\sigma_\mathrm{max}$, the coverage interval over which to place sigma-points, expressed as a number of standard deviations;
    \item $w_\mathrm{min}$, the minimum allowable weight for any sigma-point;
    \item $d_\mathrm{max}$, the maximum allowable average distance between sigma-points.
\end{itemize}
A simple grid-search is performed over these parameters, where the error relative to the Monte Carlo collision-checker is measured on the aforementioned dataset, using a simple absolute difference in collision probabilities. The latency is also measured offline on an Intel Xeon Gold 6226R Processor with a base frequency of 2.9 GHz. Figure \ref{fig:runtime_vs_error} plots the 95th-percentile of the runtime against the collision-probability error, for each set of parameter values. Table \ref{tab:runtime_metrics} shows the statistics for the chosen optimal set.
\begin{table}[htbp]
    \centering
    \caption{Runtime and latency metrics for chosen parameters: $\sigma_\mathrm{max}=3.8,\; w_\mathrm{min}=0.01,\;d_\mathrm{max}=1.625 \mathrm{m}$.}
    \begin{tabular}{lrrrr}
        \toprule
        \textbf{Metric} & \textbf{Mean} & \textbf{Median} & \textbf{P95} & \textbf{P99} \\
        \midrule
        Runtime (ms) & 0.398 & 0.213 & 0.869 & 5.215 \\
        Error (\%) &  4.1 & 3.5 & 9.3 & 11.8 \\
        \bottomrule
    \end{tabular}
    \label{tab:runtime_metrics}
\end{table}

\section{Conclusion}
\label{sec:conclusion}
This paper has presented a novel approach to collision probability estimation that addresses the key challenges of temporal correlation and computational efficiency. Through the proposed adaptive sigma-point sampling, as well as association of standardized vectors through time, the proposed algorithm is capable of accurately estimating the collision probability, while also being appropriate for real-time use.

Experimental results demonstrate consistent sub-millisecond computation times across diverse scenarios, with error rates consistently below 10\% when compared to the Monte Carlo solution. In rare cases, the computation time can approach the 5-10ms range, which is a current drawback of the approach. However, further optimizations are possible, and future work can consider addressing the ``long tail'' of the runtime distribution. 
% \section{Final copy}

% You must include your signed IEEE copyright release form when you submit your finished paper.
% We MUST have this form before your paper can be published in the proceedings.

% Please direct any questions to the production editor in charge of these proceedings at the IEEE Computer Society Press:
% \url{https://www.computer.org/about/contact}.
{
    \small
    \bibliographystyle{ieeenat_fullname}
    \bibliography{main}
}


\end{document}